\documentclass[11pt]{article}

\usepackage[final]{acl}

\usepackage{times}
\usepackage{latexsym}

\usepackage[T1]{fontenc}

\usepackage[utf8]{inputenc}

\usepackage{microtype}

\usepackage{inconsolata}

\usepackage{graphicx}

\usepackage{booktabs}
\usepackage{amsmath}
\usepackage{multirow}
\usepackage{amssymb}
\usepackage{mathtools}
\usepackage{subcaption}

\usepackage{fancyvrb}
\usepackage{fvextra}

\usepackage{xcolor}

\newcommand{\notestar}{\textsuperscript{*}}
\newcommand{\notesword}{\textsuperscript{\textdagger}}
\DeclareMathOperator*{\E}{\mathbb{E}}
\newcommand{\RPmini}{%
  \ensuremath{%
    \mathrm{RP}%
    \kern-0.1em%
    \raisebox{-0.0ex}{\scriptsize @\kern-0.15em10}%
  }%
}
\usepackage{pifont}
\newcommand{\cmark}{\ding{51}}  %
\newcommand{\xmark}{\ding{55}}  %
\newcommand{\niboldhead}[1]{\noindent\textbf{#1}} %
\newcommand{\boldhead}[1]{\textbf{#1}} %

\title{Unified Work Embeddings: \\Contrastive Learning of a Bidirectional Multi-task Ranker}

\author{
 \textbf{Matthias De Lange},
 \textbf{Jens-Joris Decorte},
 \textbf{Jeroen Van Hautte}
\\
 Techwolf
}

\begin{document}

\maketitle

\begin{abstract}
Applications in labor market intelligence demand specialized NLP systems for a wide range of tasks, 
characterized by extreme multi-label target spaces, strict latency constraints, and multiple text modalities such as skills and job titles.
These constraints have led to isolated, task-specific developments in the field, with models and benchmarks focused on single prediction tasks.
Exploiting the shared structure of work-related data,
we propose a unifying framework, combining a wide range of tasks in a multi-task ranking benchmark, and a flexible architecture tackling text-driven work tasks with a single model.
The benchmark, WorkBench, is the first unified evaluation suite spanning six work-related tasks formulated explicitly as ranking problems, curated from real-world ontologies and human-annotated resources.
WorkBench enables cross-task analysis, where we find significant positive cross-task transfer.
This insight leads to Unified Work Embeddings (UWE), a task-agnostic bi-encoder that exploits our training-data structure with a many-to-many InfoNCE objective, and leverages token-level embeddings with task-agnostic soft late interaction.
UWE demonstrates zero-shot ranking performance on unseen target spaces in the work domain, and enables low-latency inference with two orders of magnitude fewer parameters than best-performing generalist models  (Qwen3-8B), with $+4.4$ MAP improvement.
\looseness -1

\end{abstract}

\section{Introduction}
\label{sec_intro}
Applications in labor market intelligence (LMI) demand specialized NLP systems capable of scaling across diverse tasks and extreme multi-label target spaces, while meeting strict latency constraints for processing unstructured data from heterogeneous sources.
These requirements have driven isolated, task-specific developments in the field, with models and benchmarks targeting individual prediction tasks such as skill extraction or job normalization~\cite{zhang-etal-2022-skillspan,jobbert,decorte2025contextmatch_jobbertv2}.
This fragmentation poses both practical and scientific limitations, accumulating inference and maintenance costs over tasks, while precluding potential cross-task knowledge transfer and holistic progress measurement.

Concurrently, general-purpose text embedding models have advanced rapidly, demonstrating with comprehensive benchmarks that a single model can be effective across many standard NLP tasks~\cite{thakur2021beir, muennighoff2022mteb}.
However, both evaluation and models do not account for the specifics of the work domain, 
including grounding in domain-specific ontologies, extreme label cardinality, and high-speed inference.
This raises a key question: can a single, efficient embedding model replace multiple task-specific systems for real-world LMI applications?

To systematically investigate this question, we introduce WorkBench, a unified evaluation suite spanning six real-world job and skill understanding tasks. 
WorkBench includes cross-prediction tasks (predicting skills for jobs and vice versa), normalization tasks (job title and skill normalization), semantic skill search, and skill extraction, all curated from real-world sources.
All tasks are formalized as ranking problems with standardized evaluation metrics, enabling principled multi-task progress measurement.

\begin{figure*}[!ht]
  \centering
    \includegraphics[width=1.0\textwidth,trim=0pt 20pt 5pt 28pt,clip]{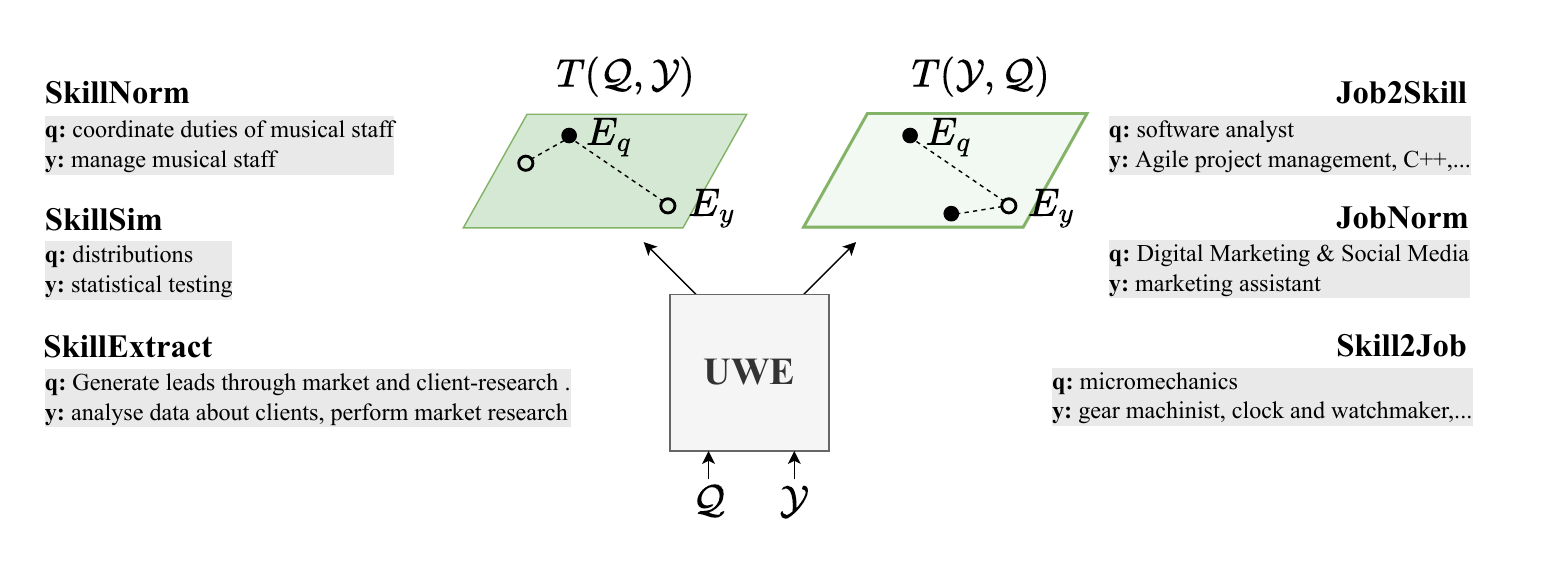}
    \vspace{-0.6cm}
  \caption{Overview of the six WorkBench tasks, demonstrating samples $q \in \mathcal{Q}$ and $y \in \mathcal{Y}$ from the query and target input spaces respectively. Unified Work Embeddings (UWE) independently encodes samples to create embeddings $E_q$ and $E_y$, and maintains a bi-directional ranking structure based on embedding similarity, enabling both tasks $T(\mathcal{Q,Y})$ and $T(\mathcal{Y,Q})$. The query-target order-agnostic setup facilitates support for all WorkBench tasks, including zero-shot performance for unseen target spaces.
  }
  \label{fig_main_overview}
\end{figure*}

Our analysis on WorkBench reveals that large general embedding models exhibit a substantial performance gap compared to domain-specific models.
Crucially, we observe significant positive transfer when training across tasks, indicating that multi-task unification is not only practical but beneficial.

Guided by these findings, we develop Unified Work Embeddings (UWE), a bidirectional multi-task embedding model built on the MPNet backbone of 109M parameters that tackles all six tasks with a shared representation space and enables real-time production deployment through embedding caching.
To address training data scarcity, we construct a scalable data pipeline producing bipartite graphs between work-related input spaces, grounded in real-world job vacancies with synthetic minority enrichment.
To learn from this graph structure, we adapt InfoNCE to the multi-task, bipartite setting with many-to-many aggregation across in-batch positive and negative relations.
Finally, we implement a SoftMax-based late-interaction module that unifies prior token-level retrieval and skill extraction architectures into a single task-agnostic component.

UWE achieves state-of-the-art results, outperforming the strongest specialist baseline (ContextMatch) by $+2.6$ MAP and $+4.0$ RP@10 macro-averaged across tasks.
Compared to generalist embedding models, UWE surpasses Qwen3-8B by $+4.4$ MAP while using $\times$73 fewer parameters and enabling $\times$3.7 lower inference latency.
These results demonstrate that a single efficient embedding model can replace multiple task-specific systems for production LMI applications, marking a significant step toward unified, deployable models for the work domain.

\section{Related Work}
\label{sec_related_work}

\boldhead{Evaluation of work-domain tasks } is challenging as high-quality structured and labeled data are scarce, while large-scale unstructured data such as business data and job postings are abundant.
Grounding unstructured data is key towards enabling real-world applications, by mapping noisy input data towards a unified and structured ground-truth ontology. 
A paramount example of such an ontology is the multilingual classification of European Skills, Competences, and Occupations (ESCO)~\cite{le2014esco}, driving many downstream evaluations due to its extensive and hierarchical label space of 3k job titles\footnote{ESCO formally distinguishes \emph{occupations} (groupings of similar jobs) from \emph{job titles} (specific labels in vacancies). We define related tasks agnostic to this difference in terminology, using ``job title'' as a unifying term for both input distributions.
} and 14k skills.
Common tasks are mapping job titles and skills to canonical examples~\cite{highprecphrase,carotene,jobbert}, predicting job titles for a given skill and vice versa~\cite{gasco2025overviewtalentclef2025skill}, extracting skills from unstructured sentences~\cite{sayfullina,zhang-etal-2022-skillspan,senger-etal-2024-deep}, and retrieving semantically similar skills~\cite{decorte2024skillmatch}. 
Recent efforts have expanded the landscape with multilingual occupation linking benchmarks~\cite{retyk2024melo}, and synthetic data generation frameworks~\cite{magron-etal-2024-jobskape}.
However, as existing downstream benchmarks mainly assess task-specific models in isolation, there remains a substantial gap in holistic evaluation and methodologies within the work domain.

\noindent\boldhead{Evaluating embedding models} with benchmarks such as 
BEIR~\cite{thakur2021beir} and MTEB~\cite{muennighoff2022mteb} has driven progress in both information retrieval and generalist embedding models. While cross-encoders excel at semantic tasks~\cite{devlin2019bert}, their computational overhead motivates bi-encoders that encode independently~\cite{reimers2019sentence}, or a combination of both with late interaction of the encoders~\cite{khattab2020colbert}.
In the work domain, however, evaluation of embedding model architectures and generalist models remains unexplored.

\section{WorkBench}\label{sec_workbench}
\subsection{Tasks}
WorkBench comprises six real-world work-related tasks in English for holistic evaluation in the work domain. Each task $T(\mathcal{Q}, \mathcal{Y}) = R$ maps query space $\mathcal{Q}$ to target space $\mathcal{Y}$, yielding ranking matrix $R \in \mathbb{R}^{|\mathcal{Q}| \times |\mathcal{Y}|}$. This uniform formalization enables multi-task evaluation with the common metrics defined in Section~\ref{sec:multitask_eval_metrics}. 
Task examples and properties are provided in Figure~\ref{fig_main_overview} and Table~\ref{tab_task_overview} respectively, with each task detailed below.

\niboldhead{Job2Skill} with test set ${T\!\left(\mathcal{J}_\text{ESCO}, \mathcal{S}_{ESCO}\right)}$ exploits the widely used ESCO ontology (v1.2.0)~\cite{le2014esco}, linking $3,039$ unique job titles and $13,939$ skills for job title space $\mathcal{J}_\text{ESCO}$ and skill space $\mathcal{S}_{ESCO}$ respectively.
This is a multi-label ranking task, as each job requires multiple skills.
For validation, we rely on a held-out subset of the training data, based on vacancy job titles, enriched by tagged ESCO skills in the vacancy description.

\niboldhead{Skill2Job} is defined as the inverted Job2Skill task ${T\!\left( \mathcal{S}_{ESCO}, \mathcal{J}_\text{ESCO}\right)}$ for both test and validation sets, and is a multi-label task with skills occurring in multiple job profiles.
\looseness=-1

\niboldhead{SkillNorm} with ${T\!\left(\mathcal{A}_\text{ESCO}, \mathcal{S}_{ESCO}\right)}$ exploits ESCO's alternative label structure where each skill in $\mathcal{S}_{ESCO}$ is provided with several alternative formulations, which we define as $\mathcal{A}_\text{ESCO}$. 
Defining the task to predict the canonical skill for each alternative enables a fixed target output space independent of the number of alternatives.

\niboldhead{SkillExtract} relies on the SkillSpan~\cite{zhang-etal-2022-skillspan} \emph{House} and \emph{Tech} subsets, enriched by subsequent work with fine-grained ESCO (v1.1.0) skill labels per span~\cite{decorte2022design}. We report the subtasks separately, denoted as SkillExtr-H and SkillExtr-T respectively. The task is to predict from job vacancy sentences $\mathcal{V}$ the set of skills it contains ${T\!\left(\mathcal{V}, \mathcal{S}_{ESCO}\right)}$.

\niboldhead{SkillSim} is a semantic skill similarity task $T(\mathcal{S}_{q}, \mathcal{S}_{q \, \cup \,y})$ based on the thousand related skill pairs in SkillMatch-1k~\cite{decorte2024skillmatch}. As it comprises a single test dataset, we make a 10/90 split into validation and test sets.
While using the original query skills $\mathcal{S}_q$, we define an extended target space based on all $2,648$ unique query and target skills $\mathcal{S}_{q \, \cup \,y}$in the dataset, but excluding each query itself in the resulting ranking scores.

\niboldhead{JobNorm} relies on the open-sourced dataset from JobBERT~\cite{jobbert}, containing governmental job vacancy titles $\mathcal{J}$ that are tagged with corresponding ESCO (v1.0.5) job titles $\mathcal{J}_\text{ESCO}$, resulting in task ${T\!\left(\mathcal{J}, \mathcal{J}_\text{ESCO}\right)}$. 
We use ESCO's canonical preferred labels as target space to maintain comparability with prior work.

\begin{table}[ht]
  \centering
  \small
  
  \resizebox{\columnwidth}{!}{
    \begin{tabular}{@{}l c c c c@{}}
      \toprule
      \textbf{Task} & \textbf{Label} & \textbf{Val} & \textbf{Test} & $|\mathcal{Y}|$ \\
      \midrule
      Job2Skill            & multi   & 5,644\notesword  & 3,039\notestar  & 13,939            \\
      Skill2Job           & multi   & 10,137\notesword & 13,492\notestar & 5,644 / 3,039              \\
      SkillNorm            & one  & 13,222\notestar & 73,311\notestar & 13,939            \\
      SkillExtr-T  & multi   & 75\notesword     & 338\notesword    & 13,891            \\
      SkillExtr-H & multi   & 61\notesword     & 262\notesword    & 13,891            \\
      SkillSim             & one  & 100              & 900              & 2,648             \\
      JobNorm              & one  & 15,462\notesword & 15,461\notesword & 2,942             \\ %
      \bottomrule
    \end{tabular}
  }
    \caption{\label{tab_task_overview}Overview of WorkBench task size for validation (val) and test set (test), ranking's target space $\mathcal{Y}$, and multi-hot or one-hot labels. ‘†’ indicates only one of target or query space is ESCO-based, and ‘*’ indicates that both are.}
\end{table}

\subsection{Multi-task Evaluation}
\label{sec:multitask_eval_metrics}
WorkBench relies on two complementary information retrieval metrics. Mean Average Precision (MAP) averages the precision at each relevant item across the full ranking, capturing both the order and completeness of relevant predictions; for single-label tasks it reduces to the Mean Reciprocal Rank. R-Precision@10 (\RPmini{}) measures the fraction of relevant items among the top 10 results, providing a more application-oriented view of early retrieval quality~\cite{metrics_ap_mrr}.
All task types are allocated equal importance in the final multi-task metric. Therefore, metric results are first averaged per task type (in particular subsets SkillExtr-H and SkillExtr-T), and for each metric we report the final macro average over all tasks.

\begin{figure*}[!ht]
  \centering
    \begin{subfigure}[t]{0.32\textwidth}
    \centering
    \includegraphics[width=\linewidth,
                     trim=14pt -15pt 10pt 10pt, clip]{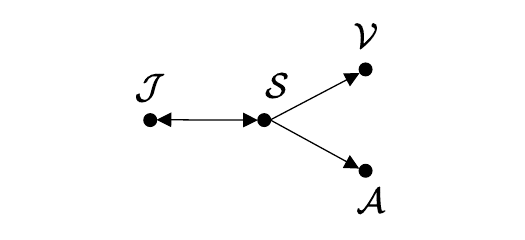}
    \caption{\label{fig_data_process_deps}}
  \end{subfigure}
  \begin{subfigure}[t]{0.33\textwidth}   %
    \centering
    \includegraphics[width=\linewidth,
                     trim=22pt 0pt 25pt 11pt, clip]{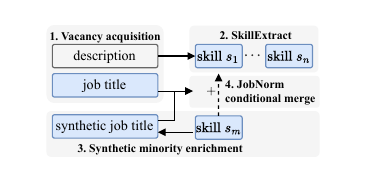}
    \caption{\label{fig_data_pipeline}}
  \end{subfigure}\hfill
  \begin{subfigure}[t]{0.32\textwidth}
    \centering
    \includegraphics[width=\linewidth,
                     trim=7pt 7pt 7pt 7pt, clip]{img/skill_distr_improved}
    \caption{\label{fig_skill_distr}}
  \end{subfigure}\hfill
  \vspace{-0.1cm}
  \caption{\label{fig_data_process}
  Overview of the training data.
  (a) Data dependency graph of the training data, showing the many-to-many and one-to-many edges from skill space $\mathcal{S}$ to job title $\mathcal{J}$, vacancy sentences $\mathcal{V}$, and skill alternatives $\mathcal{A}$ spaces.
   (b) Acquisition overview of structured training data with job titles and grounded skills. Blue colored boxes indicate structured fields.
    (c) Log-frequencies for the skills in the raw vacancy training dataset, before and after synthetic enrichment, with highlighted background in red indicating the gap of represented skills. 
  }
\end{figure*}

\section{Method\label{sec_method}}

\subsection{Task-agnostic encoding}
The task diversity in WorkBench imposes a significant first challenge on the flexibility of the model architecture. 
Secondly, not all tasks have labeled training data available, making zero-shot inference with unseen text labels a hard requirement. 
Thirdly, WorkBench's label cardinality $|\mathcal{Y}|$ reaches up to 14k (see Table~\ref{tab_task_overview}), making pair-wise inference infeasible for low-latency production settings.

While cross-encoders can achieve improved performance with pair-wise inference~\cite{reimers2019sentence}, they scale linearly with $|\mathcal{Y}|$.
In contrast, bi-encoders process textual inputs independently to create a metric space, and have shown state-of-the-art performance on job normalization and skill extraction~\cite{decorte2025contextmatch_jobbertv2}.
However, the bi-encoder training and architecture remain tailored to a single task. 
Therefore, we propose Unified Work Embeddings (UWE), a task-agnostic bi-encoder that generates embeddings for any query $q\in\mathcal{Q}$ and target $y\in\mathcal{Y}$ in WorkBench, effectively enabling a bi-directional ranking structure based on embedding similarity, supporting both tasks $T(\mathcal{Q,Y})$ and $T(\mathcal{Y,Q})$. At inference for a given task, only a single forward pass is required for $q$, as the model can rely on a target space embedding cache. 
In the following, we detail the process of acquiring graph-structured training data, the loss objective used to exploit this structure data, and an architectural shift to soft late interaction of the embeddings.

\subsection{Obtaining Structured Training data}
To obtain a unified embedding model, it's crucial to obtain representative training data for the input domains. From Section~\ref{sec_workbench} we identify a skill space $\mathcal{S}$, job title space $\mathcal{J}$, skill alternatives $\mathcal{A}$, and vacancy sentences $\mathcal{V}$.
Following common practice, a common skill ontology is made available to the model~\cite{malandri2025skillmo,decorte2025contextmatch_jobbertv2}. We ground the data in the ESCO skill vocabulary, and exploit its relations with the remaining input spaces to create an interconnected data structure, as exemplified in Figure~\ref{fig_data_process_deps}.

\boldhead{Real-world data extraction.} Capturing a representative real-world distribution of job titles, we use scraped  job vacancies from a public US jobboard, with 10k monthly job postings over a year in 2023. The data is semi-structured, providing a job title and description of the vacancy. The description is postprocessed with a state-of-the-art skill extraction model~\cite{decorte2025contextmatch_jobbertv2} to obtain a set of ESCO skill predictions.  This results in 104k vacancies with at least one confident skill prediction, covering 3.5M skill predictions with on average 33 skills per vacancy.
A crucial step to ensure data diversity is deduplication and aggregation of the job titles. After case-agnostic merging of job titles, 57k unique job titles remain. The skills of matched duplicate titles are merged through majority voting based on occurrence in the duplicates. Excessive skill profiles are cut off on the top-200, ordered on averaged prediction confidence.
With a 90/10 split we obtain the train and validation sets, with validation set being used for Job2Skill and Skill2Job.  Figure~\ref{fig_data_pipeline} shows an overview of the process in steps 1 and 2.

\boldhead{Inversed synthetic minority enrichment.} Analyzing the resulting skill distribution in Figure~\ref{fig_skill_distr}, it shows to be extremely long-tailed. General skills such as "communication" have 16k occurrences, while many other skills are underrepresented, including $862$ non-present ESCO skills such as "harvest grapes" and "understand written Urdu".
To this end, we reverse the data collection by grounding it in the ESCO skills, and using a generative model (GPT4o-mini) to generate a set of minimum 20 job titles that would require the skill. 
However, the resulting 285k job titles have sparse skill profiles with just a single skill rather than a many-to-many structure.
A first approach to get one-to-many relations is to deduplicate the generated job titles, but this enables merging skills from only a small fraction of exact matches.
Therefore, we rely on a state-of-the-art job normalization model~\cite{decorte2025contextmatch_jobbertv2} to find for each synthetic job title the best matching vacancy job titles. After a match, the skill from the synthetic job title is added to the existing vacancy job title's skills. This approach enables at least one enrichment of rare skills for all vacancy job titles, via 165k confident matches.
The remaining synthetic job titles without matches are also added to the dataset.
The synthetic enrichment grows the dataset from 1.8M to 3.3M job-skill pairs. Figure~\ref{fig_data_pipeline} depicts the process in steps 3 and 4.

\boldhead{Synthetic vacancy training data.}
For the vacancy sentences, we rely on the synthetic dataset from prior work~\cite{decortellm} that shows increased efficacy with 10 synthetically generated ad sentences per ESCO skill (v1.1.0).
The dataset has $97\%$ coverage for the ESCO v1.2.0 skills used in our grounding. We add the official ESCO description for all skills to ensure full coverage.

\boldhead{Synthetic skill alias training data.}
Similarly, we exploit controlled generation of synthetic data using GPT4o-mini~\cite{achiam2023gpt} for alternative skill formulations. The model context is provided with three representative examples from the SkillNorm validation set, to generate 20 alternatives per ESCO skill.
Full details are provided in Appendix~\ref{apdx:data_pipeline_details}, with contributions of synthetic components quantified in the ablation study in Section~\ref{sec:ablation_study}.

\subsection{Many-to-many InfoNCE over bipartite graphs}
Designing our multi-task training data structure, we find a highly interconnected structure as depicted in Figure~\ref{fig_data_process_deps}, where skills exhibit many-to-many edges with the job titles, and one-to-many edges with vacancy sentences and skill variants. 
As we will show in the following, traditional InfoNCE objectives fall short in this scenario. 

Assume a mini-batch $B$ is constructed as a sequence of query-target pairs $(q, y_+)$, where the query is a skill $q\in \mathcal{S}$ and the target is a matching positive from any of the remaining spaces $ y_+ \in \mathcal{J}\cup \mathcal{V}\cup\mathcal{A}$. Let $Y$ be the set of all targets in the batch, then the InfoNCE loss~\cite{oord2018infonce} for query $q$ is 
\begin{equation}
\mathcal{L}_{\text{InfoNCE},q}
=-
    \log
    \frac{
        \exp\bigl( \texttt{sim} ( q, y_+ ) / \tau\bigr)
    }{
        \displaystyle\sum_{y\in Y}
        \exp\bigl( \texttt{sim} ( q, y ) / \tau\bigr)
    }
,
\label{eq_infonce}
\end{equation}
where $\texttt{sim} ( q, y) = f_\theta(q) \cdot f_\theta(y)$ denotes dot product similarity of the samples encoded by model $f_\theta$, and $\tau$ is the scaling temperature.

The InfoNCE loss exhibits three main problems. 
First, the loss considers pairs independently and disregards cross-pair relations, as all other-pair targets are regarded as in-batch negatives. A skill that occurs in the batch with a job title $y_+$, will regard any remaining occurrences of the skill with other targets as negatives in $Y$.
Second, the loss is unaware of the multi-task data structure, entangling the contribution to the loss from all tasks. This makes the weighing explicitly prone to imbalanced task distributions. 
Thirdly, the loss is asymmetric, and therefore unaware of many-to-many relations. Not only can a skill be related to multiple job titles, but a target job title $y$ can also have multiple skills.
To address these issues, we introduce a contrastive many-to-many  (MTM) InfoNCE loss.
To decompose the interconnected dependency graph, we rely on the centrality of the skill space $\mathcal{S}$ in our training data structure. This enables a decomposition in bipartite graphs for each of $\mathcal{J, V, A}$ with $\mathcal{S}$.

As a first step, we define an asymmetric loss to account for one-to-many relations in a single bipartite graph.
Given a mini‑batch of $N$ query-target candidate pairs, we define bipartite graph
$G \;=\; (Q \cup Y,\,D)$,
where
\begin{itemize}
  \item $Q=\{\,q_1,\dots,q_{N} \,\} \subseteq \mathcal{Q}$ is the set of \emph{query nodes} in the batch, each represented by its textual input $q_i$;
  \item $Y=\{\,y_1,\dots,y_{N}\,\} \subseteq \mathcal{Y}$ is the set of batch \emph{target nodes}, each with an input text $y_i$;
  \item $D\subseteq Q\times Y$ is the set of observed edges, where an edge $(q,y)\in D$ indicates that query $q$ is associated with target~$y$.
\end{itemize}

Let $Y_{q}\;=\;\bigl\{\,y\in Y \;\big\vert\; (q,y)\in D\bigr\}$
denote the \emph{target set} for query $q$.  
By construction, we ensure $\lvert Y_{q}\rvert \ge 1$ for every $q$.
In the context of self-supervised learning with images, \citet{khosla2020supcon} find that for multi-label targets, averaging the InfoNCE yields superior performance by averaging outside of the log-function. We exploit this insight to enable node-wise averaging over the queries and over the query target set as
\begin{equation}
\mathcal{L}_{Y\mid Q}
=-
\E_{{\substack{q\sim Q \\  y_+\sim Y_{q}}}}
    \log
    \frac{
        \exp\bigl( \texttt{sim} ( q, y_+ ) / \tau\bigr)
    }{
        \displaystyle\sum_{y\in Y}
        \exp\bigl( \texttt{sim} ( q, y ) / \tau\bigr)
    }
.
\label{eq_asym_loss}
\end{equation}
While this is a one-to-many objective over each of the nodes in $Q$, the same can be applied for the target nodes $Y$, hence, creating the contrastive many-to-many objective as:
\begin{equation}
\mathcal{L}_{Y, Q} = \mathcal{L}_{Y\mid Q} + \mathcal{L}_{Q\mid Y}.
\label{eq_sym_loss}
\end{equation}

Having a loss for a single bipartite graph, the final loss is the weighted sum over all bipartite graphs in our data structure:
\begin{equation}
    \mathcal{L}_\text{MTM} = \alpha_J \mathcal{L}_{S, J} + \alpha_V\mathcal{L}_{S, V} + \alpha_A \mathcal{L}_{S, A}
    \label{eq_mtm}
\end{equation}%
where instead of a single skill-centered graph, the separability in bipartite graphs allows controlled importance weighing with $(\alpha_J,\alpha_V,\alpha_A)$ over subsets of the multi-task data structure.

\begin{table*}[!th]
\centering
\resizebox{\textwidth}{!}{%
\setlength{\tabcolsep}{5pt}%
\begin{tabular}{@{}lllllllllllllllll@{}}
\toprule
                      & \multicolumn{2}{c}{\textbf{Job2Skill}} & \multicolumn{2}{c}{\textbf{Skill2Jobs}} & \multicolumn{2}{c}{\textbf{JobNorm}} & \multicolumn{2}{c}{\textbf{SkillNorm}} & \multicolumn{2}{c}{\textbf{SkillSim}} & \multicolumn{2}{c}{\textbf{SkillExtr-T}} & \multicolumn{2}{c}{\textbf{SkillExtr-H}} & \multicolumn{2}{c}{\textbf{Task Avg}} \\
             & MAP      & \RPmini     & MAP      & \RPmini      & MAP     & \RPmini    & MAP      & \RPmini     & MAP     & \RPmini     & MAP       & \RPmini      & MAP       & \RPmini      & MAP     & \RPmini     \\ \midrule

\textbf{UWE (Ours)} & \textbf{17.9} & \textbf{37.8} &\textbf{ 37.3} & \textbf{49.4} & \textbf{39.3} & \textbf{60.8} & \textbf{88.9} & \textbf{95.9} & 10.4 & 33.8 & 52.3 & 73.5 & \textbf{42.3} & \multicolumn{1}{l|}{64.2} & \textbf{40.2} & \textbf{57.7} \\

\addlinespace\emph{Task-specific models}\\
\textbf{ContextMatch} & 14.6            & 34.5                 & 32.2             & 43.3                 & 36.8           & 55.3                & 86.5            & 94.3                 & 7.7             & 24.6                & \textbf{53.6}    & \textbf{73.9}         & 42.2          & \multicolumn{1}{l|}{\textbf{65.8}} & 37.6            & 53.7                \\
\textbf{JobBERT-v2}   & 0.4             & 0.3                  & 0.5              & 0.4                  & 39.0  & 58.5                & 83.0            & 91.8                 & 9.7             & 30.4                & 35.1             & 48.5                  & 20.8          & \multicolumn{1}{l|}{38.1}          & 26.8            & 37.4                \\
\addlinespace\emph{Gen. emb. models}\\
\textbf{MPNet}        & 11.8            & 29.0                 & 29.0             & 40.3                 & 32.9           & 50.8                & 84.7            & 93.2                 & 10.1            & 30.7                & 31.2             & 49.8                  & 19.3          & \multicolumn{1}{l|}{36.7}          & 32.3            & 47.9                \\
\textbf{Emb. Gemma} & 12.6 & 30.1 & 29.9 & 41.6 & 36.7 & 54.6 & 85.1 & 93.9 & 9.3 & 28.7 & 38.1 & 59.3 & 27.3 &  \multicolumn{1}{l|}{44.9} & 34.4 & 50.2 \\
\textbf{E5-mistral-7B} & 11.1 & 28.7 & 29.2 & 40.4 & 35.3 & 54.4 & 83.4 & 92.9 & 12.2 & 37.7 & 32.0 & 45.8 & 19.6 & \multicolumn{1}{l|}{33.7} & 32.8 & 49.0 \\
\textbf{Qwen3-0.6B}   & 11.7            & 29.1                 & 28.3             & 39.7                 & 30.1           & 49.5                & 81.7            & 92.2                 & 10.4            & 32.7                & 35.1             & 54.6                  & 25.7          & \multicolumn{1}{l|}{41.8}          & 32.1            & 48.6                \\ 
\textbf{Qwen3-4B} & 13.0 & 31.6 & 31.9 & 43.8 & 33.6 & 53.7 & 85.1 & 94.1 & 12.3 & 37.9 & 37.6 & 58.4 & 28.9 & \multicolumn{1}{l|}{46.4} & 34.8 & 52.3 \\
\textbf{Qwen3-8B} & 13.8 & 32.7 & 32.7 & 44.4 & 35.3 & 55.5 & 85.7 & 94.5 & \textbf{13.0} & \textbf{40.8} & 39.1 & 61.0 & 29.1 & \multicolumn{1}{l|}{49.0} & 35.8 & 53.8 \\
\bottomrule
\end{tabular}%
}
\caption{
WorkBench results (MAP and RP@10, \%) for UWE, task-specific, and generalist embedding models. Task Avg is the macro-average over all tasks. Best result per column in \textbf{bold}.
\label{tab_main_workbench_results}
}
\end{table*}

\subsection{Task‑Agnostic Soft Late Interaction}
\label{subsec_tasli}

Late‑interaction architectures such as ColBERT~\cite{khattab2020colbert} narrow the performance gap between bi‑encoders and cross‑encoders by enabling token-level similarity computation after independent encoding of query and document tokens. While ColBERT focuses on sparse interaction by only looking at maximally similar token pairs, ContextMatch~\cite{decorte2025contextmatch_jobbertv2} has demonstrated that a softmax-based alternative with token‑level information on the sentence-side markedly improves skill extraction. 
However, the latter is task-specific to skill extraction, while we require a holistic architecture to support all WorkBench tasks.
To this end, we jointly study late interaction in retrieval and task‑specific extraction and generalise it into a single, task‑agnostic module with two-sided token-level interaction and SoftMax-based fusion.

Given a query $q$ and target $y$, let $\langle t_{1},\dots,t_{n}\rangle$ and $\langle u_{1},\dots,u_{m}\rangle$ be their respective token sequences encoded by a shared encoder $f_\theta$ into matrices $E_q\in\mathbb{R}^{n\times h}$ and $E_y\in\mathbb{R}^{m\times h}$ of non-normalized embeddings, where $h$ is the embedding dimensionality.
Their pairwise token‑similarity is defined by 
$S =  E_q  E_y^{T}\in\mathbb{R}^{n\times m}$, and $\hat S =  \hat E_q  \hat E_y^{T}$ denotes token-similarity for row-wise L2-normalized embeddings $\hat E_q, \hat E_y$.
Subsequently, the interaction matrix $A$ defines the importance for each of the token-interactions. Instead of a sparse binary matrix with \textsc{MaxSim} pooling as in ColBERT, we employ a temperature‑controlled row-wise softmax attention:
\[
A \;=\; \operatorname{SoftMax}\bigl(S/\tau_a\bigr), \qquad
\textstyle\tau_a>0,
\]
followed by a Frobenius inner product
\[
\operatorname{sim}(q,y)
\;=\;
\langle A , \hat S \rangle_F .
\]
This \emph{soft} aggregation supplies two benefits:
(i) it yields a gradient signal for all token‑to‑token interactions with controllable distribution smoothing, converging to \textsc{MaxSim} for $\tau_a\!\to\!0$
, and
(ii) it is task-agnostic as it abstains from the skill extraction tailored mechanism in ContextMatch that applies mean-pooling on $E_y$ in the skill space.

\section{WorkBench results}\label{sec_workbench_results}

\subsection{Baselines}
We compare against state-of-the-art task-specific and general-purpose baselines.
\textbf{JobBERT-v2} and \textbf{ContextMatch} are two bi-encoders tailored to job normalization and skill extraction respectively~\cite{decorte2025contextmatch_jobbertv2}. Both are based on a general pretrained \textbf{MPNet} backbone of 109M parameters~\cite{song2020mpnet}
\footnote{\scriptsize
    \url{https://huggingface.co/sentence-transformers/all-mpnet-base-v2}
  }%
that we also include as baseline representing a domain-agnostic embedding model.
Further, we obtain the state-of-the-art general embedding models over a range of model capacities on MTEB~\cite{muennighoff2022mteb}, where \textbf{Qwen3} (0.6B, 4B, 8B)~\cite{qwen3embedding}, \textbf{E5-mistral-7B}~\cite{wang2023e5mistral}, and \textbf{EmbeddingGemma}~\cite{vera2025embeddinggemma} are on top of the leaderboard, and their open weights enable reproducibility of our experiments. Table~\ref{tab_latencies} summarizes model capacity and embedding-dimensionality.

\subsection{Experimental setup \label{sec_setup_details}}
The training pipeline for UWE is grounded in the known set of ESCO skills through uniform sampling, resulting in zero-shot predictions for Skill2Job, SkillSim, and JobNorm, as detailed in Section~\ref{sec:zero_shot}. To enable comparison with the domain expert models, UWE exploits the MPNet backbone model~\cite{song2020mpnet}.
Training length is defined on the largest number of unique edges in either of the $\mathcal{J, V, A}$ spaces, which are the $3.3$M job-skill pairs.
For a batch of $512$ skills, $512$ samples for each of the $\mathcal{J, V, A}$ spaces are sampled, resulting in a total $2,048$ input texts. Following \citet{decorte2025contextmatch_jobbertv2}, we augment vacancy sentences with a random sample in $\mathcal{V}$ as prefix or suffix. 
We use a linear learning rate schedule, with factor $0.1$ warm-up and peak learning rate of $8e^{-5}$.
The WorkBench validation sets are used to measure task-average performance every 50 updates, from which the best-performing checkpoint on the main MAP metric is preserved.
For latency and floating-point operations (FLOPS), we benchmark each model in isolation, measuring the total cost of both encoding a single query and computing similarity scores against pre-cached target embeddings (bf16 precision, batch size 1). We report the mean and standard error over all WorkBench tasks.
All experiments are run on a NVIDIA A100-SXM4-40GB GPU. For a full overview of configuration and optimized hyperparameters, we refer to Appendix~\ref{apdx:exp_details}.

\subsection{Results and Discussion}

Table~\ref{tab_main_workbench_results} reports UWE and all baselines across WorkBench tasks.
Averaged over all task types, UWE outperforms the best-performing task-specific model ContextMatch with $+2.6$ MAP and $+4.0$ RP@10, and the best-performing general embedding model Qwen3-8B with $+4.4$ MAP and $+3.9$ RP@10.
Looking at task-specific results, UWE is state-of-the-art on Job2Skill, Skill2Jobs, and SkillNorm.
For the JobNorm task, UWE outperforms the task-tailored model JobBERT-v2 in terms of MAP and achieves $+2.3$ on RP@10.
ContextMatch retains state-of-the-art results on SkillExtr-T, but is outperformed by UWE in MAP on SkillExtr-H.
ContextMatch has the second-best average performance over tasks, indicating that the skill extraction task entails positive cross-task transfer, as we confirm in Section~\ref{sec_task_transfer}.
The general embedding models perform increasingly well on SkillSim with increasing model capacity. 
Nonetheless, UWE performs on par with Qwen3-0.6B while having a factor $5.5$ fewer parameters. Increasing Qwen3's model capacity to 4B and 8B parameters leads to $12.3$ and $13.0$ MAP. 
Inspecting Qwen's results on the capacity-accuracy Pareto frontier on MTEB~\cite{muennighoff2022mteb,qwen3embedding}, Qwen3's results increase from an average $64.34$ for $0.6$B parameters to $+5.11$ ($4$B) and $+6.24$ (8B). Our results show that the same trend holds for WorkBench with  $+2.7$ ($4$B) and $+3.7$ (8B) over the $0.6$B model.

Table~\ref{tab_latencies} reports for all models the capacity and embedding-dimensionality, next to the task-averaged latency and FLOPS per query in WorkBench.
UWE outperforms all models on average ranking performance, while having nearly two orders of magnitude fewer parameters than Qwen3-8B.
MPNet-based models UWE, ContextMatch, and JobBERT-v2, all show the lowest latency and FLOPS/query. While token-level late interaction shows increased FLOPS/query with $1.57 \pm 0.35$B for ContextMatch, and $3.01 \pm 0.94$B for UWE, 
both methods preserve similar real-time latency.
UWE and the other MPNet-based models retain at least a factor $\times 3$ lower latency compared to all generalist baselines.

\begin{table}[!th]
\centering
\resizebox{\linewidth}{!}{%
\begin{tabular}{@{}lllll@{}}
\toprule
\textbf{Model}        & \textbf{Param.} & \textbf{$h$} & \textbf{Latency (ms)} & \textbf{FLOPS} \\ 
\midrule
\textbf{UWE (Ours)}   & 109M $(\times1)$   & 768          & $15.9 \pm 0.6$             & $3.0 \pm 0.9$B     \\
\textbf{ContextMatch} & 109M               & 768          & $16.1 \pm 0.7$             & $1.6 \pm 0.4$B     \\
\textbf{JobBERT-v2}   & 109M               & 1,024        & $13.4 \pm 0.2$             & $1.3 \pm 0.2$B     \\
\addlinespace
\textbf{MPNet}        & 109M               & 768          & $13.2 \pm 0.2$             & $1.3 \pm 0.2$B     \\
\textbf{Emb. Gemma} & 308M  $(\times2.8)$             & 768          & $58.9 \pm 0.1$             & $1.5 \pm 0.3$B     \\
\textbf{E5-mistral-7B} & 7B  $(\times65.2)$             & 4,096          & $61.6 \pm 8.6$             & $112.0 \pm 21.0$B     \\
\textbf{Qwen3-0.6B}   & 596M $(\times5.5)$ & 1,024        & $46.2 \pm 0.4$             & $5.9 \pm 1.2$B     \\
\textbf{Qwen3-4B}     & 4B $(\times36.7)$  & 2,560        & $60.0 \pm 0.1$             & $48.4 \pm 9.9$B    \\
\textbf{Qwen3-8B}     & 8B $(\times73.4)$  & 4,096        & $60.2 \pm 0.8$             & $92.4 \pm 19.0$B   \\
\bottomrule
\end{tabular}%
}
\caption{
Overview of model's number of parameters (Param.), embedding dimensionality ($h$), and the benchmarked inference latency per query and FLOPS per query to obtain a ranking, showing the mean $\pm SE$ over all WorkBench tasks.
\label{tab_latencies}
}
\end{table}

\section{Analysis}

\subsection{Cross-Task Transfer \label{sec_task_transfer}}
The three bipartite loss components in $\mathcal{L}_\text{MTM}$ match closely with specific tasks in WorkBench: 
$\mathcal{L}_{S,J}$ explicitly matches skills and jobs (Skill2Job, Job2Skill),  $\mathcal{L}_{S,V}$ focuses on vacancy sentences for skill extraction (SkillExtract), and $\mathcal{L}_{S,A}$ matches with the SkillNorm task.
To understand how each of the loss components contributes to the tasks in WorkBench, we study them in isolation and measure the cross-task knowledge gain. 
Knowledge Gain is the delta of the newly trained model's performance, compared to the base model with parameters $\theta_0$ at training initialization: $MAP_{\theta_t} - MAP_{\theta_0}$.
For this experiment, due to faster convergence, we set training to 300 steps, and to allow fair comparison with the base model that has not been trained on token-level embeddings, all models employ mean-pooling and produce a ranking based on cosine similarity. 

\begin{figure}[!t]
  \centering
    \includegraphics[width=0.9\columnwidth,trim=5pt 10pt 5pt 6pt,clip]{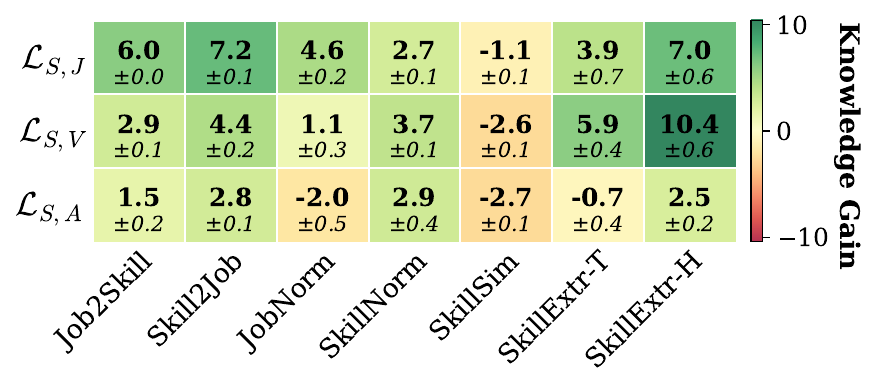}
    \caption{Task transfer experiment showing our three partial MTM-loss objectives $\mathcal{L}_{S,J}$, $\mathcal{L}_{S,V}$, $\mathcal{L}_{S,A}$ trained in isolation, starting from the base MPNet model. Knowledge Gain is the delta-increase in MAP, averaged over 5 runs ($\pm$ SE). 
  \label{fig_heatmap_task_transfer}
  }
\end{figure}

Figure~\ref{fig_heatmap_task_transfer} shows the loss components trained in isolation on the y-axis, and the WorkBench tasks on the x-axis. 
As expected, $\mathcal{L}_{S,J}$ improves in Job2Skill ($+6.0$) and Skill2Job ($+7.2$). However, we find it also contributes significant knowledge gain to JobNorm ($+4.6$), SkillNorm ($+2.7$) and both SkillExtract tasks ($+3.9, +7.0$). 
$\mathcal{L}_{S,V}$ shows a similar pattern, and improves the most on its matching SkillExtract tasks.
$\mathcal{L}_{S,A}$ improves most for alternative skills in SkillNorm ($+2.9$), has positive gain on 3 tasks, and shows negative transfer for JobNorm and SkillExtr-T.
When all three objectives are brought together, we find an increase in SkillSim performance (Table~\ref{tab_main_workbench_results}), suggesting that the combination provides a regularization effect that enables more generalizable skill representations, in contrast to the negative transfer of any individual objective.\looseness=-1

\begin{table}[!t]
\centering
\resizebox{\linewidth}{!}{%
\begin{tabular}{@{}lcccll@{}}
\toprule
 \textbf{Loss}                                    & \textbf{Sym.}         & \textbf{Multi-Label}   & \textbf{Struct.} & \textbf{MAP} \\ \midrule
$\mathcal{L}_{\text{InfoNCE}}$                              & \xmark & \xmark & \xmark &          36.40 $\pm$ 0.07 \\
$\mathcal{L}_{\text{MTM}}$-pairwise                         & \cmark & \xmark & \cmark & 38.17 $\pm$ 0.18          \\
$\mathcal{L}_{\text{MTM}}$                           & \cmark & \cmark & \cmark & \textbf{38.33 $\pm$ 0.24}  \\ \bottomrule
\end{tabular}%
}
\caption{Ablation of the MTM loss, evaluating its symmetry (sym.), multi-label support (multi-label), and structure awareness (struct.); reporting MAP (95\% CI over 5 runs).
\label{tab_loss_ablation}
}
\end{table}

\subsection{Ablation Study}
\label{sec:ablation_study}
We perform ablations for both the loss and soft late-interaction components. This study applies a training regime of 600 model updates without early stopping, averaged over 5 runs, reporting task-average MAP on WorkBench. Details can be found in Appendix~\ref{apdx:exp_details}.
For the loss component, Table~\ref{tab_loss_ablation} reports across three dimensions: asymmetric versus symmetric application of the loss in both directions (Eq.~\ref{eq_sym_loss}), per-label versus multi-label-aware aggregation over multiple positives per query (Eq.~\ref{eq_asym_loss}), and a single batch of positives and negatives versus the decomposition into task-specific bipartite graphs (Eq.~\ref{eq_mtm}).
Per Eq.~\ref{eq_infonce}, the standard $\mathcal{L}_{\text{InfoNCE}}$ ignores all three properties, yielding $36.40$ MAP. Introducing our proposed $\mathcal{L}_{\text{MTM}}$ improves performance by an average of $+1.93 $ across tasks, whereas the variant that omits multi-label relationships (denoted $\mathcal{L}_{\text{MTM}}$-pairwise) suffers a $0.16 $ drop.

For soft late interaction, Table~\ref{tab_LI_numbers} compares to sentence embeddings, the common approach of cosine similarity with mean-pooled token embeddings in both $E_q$ and $E_y$~\cite{reimers2019sentence}.
We find that token-level interaction with MaxSim~\cite{khattab2020colbert}, with a sparse and binary interaction matrix, is not guaranteed to outperform sentence embeddings on WorkBench.
In contrast, we find that both token-level SoftMax approaches yield a significant gain, where our proposed method of exposing tokens on \emph{both} sides achieves the best accuracy. 
Token-to-token interaction increases inference FLOPS by $\times$1.9, while remaining real-time with a limited increase of $\approx 2.5$ms in latency, as reported in Table~\ref{tab_latencies}.

For the data pipeline, Table~\ref{tab_synth_ablation} quantifies the contribution of synthetic data components.
Removing synthetic aliases (w/o alias synth.) reduces task-average MAP by $0.3$, while further removing all synthetic enrichment (w/o all synth.) yields a $1.7$ drop. Per-task results are reported in Appendix~\ref{apdx:synth_ablation_full}.
\looseness=-1

\begin{table}[!th]
\centering
\resizebox{\linewidth}{!}{%
\begin{tabular}{@{}lccc@{}}
\toprule
\textbf{Late Interaction}  & $E_q$-mean    &  $E_y$-mean  & \textbf{MAP}     \\ \midrule
sentence embeddings     & \cmark   &  \cmark    & 38.00 $\pm$ 0.41 \\
$\operatorname{MaxSim}$      & \xmark   &  \xmark               & 37.63 $\pm$ 0.52 \\
$\operatorname{SoftMax}$ - $y$-mean   & \xmark   &  \cmark    & 39.12 $\pm$ 0.10 \\ 
$\operatorname{SoftMax}$  & \xmark   &  \xmark & \textbf{39.45 $\pm$ 0.20} \\
\bottomrule
\end{tabular}%
}
\caption{Ablation of soft late interaction, reporting task-averaged MAP on WorkBench ($95 \%$ CI over 5 runs). \label{tab_LI_numbers}}
\end{table}

\begin{table}[!th]
\centering
\resizebox{0.55\linewidth}{!}{%
\begin{tabular}{@{}lcc@{}}
\toprule
& \textbf{MAP} & \textbf{{\boldmath\RPmini}} \\
\midrule
\textbf{UWE}        & \textbf{40.2} & \textbf{57.7} \\
w/o alias synth.     & 39.9 & 57.1 \\
w/o all synth.       & 38.5 & 55.9 \\
\bottomrule
\end{tabular}%
}
\caption{
Synthetic data ablation reporting task-averaged metrics on WorkBench, removing synthetic skill aliases (w/o alias synth.) and all synthetic enrichment (w/o all synth.) from UWE training data.
\label{tab_synth_ablation}}
\end{table}

\subsection{Zero-Shot Generalization}
\label{sec:zero_shot}

To substantiate UWE's task-agnostic claims, we evaluate zero-shot generalization at three increasingly challenging levels: (i) unseen target space, where the evaluation target distribution $\mathcal{Y}$ was not observed during training; (ii) unseen task, where no corresponding training objective exists; and (iii) unseen ontology, where both query and target originate from an entirely different taxonomy.

\niboldhead{Unseen target space.}
In Skill2Job, the target space consists of canonical ESCO occupation labels $\mathcal{J}_\text{ESCO}$, which are clean, standardized titles absent from training, where only noisy vacancy-derived job titles are observed.

\niboldhead{Unseen task.}
In \text{JobNorm}, the model must match vacancy titles to canonical occupations, a job-to-job task with no corresponding training objective.
\text{SkillSim} requires ranking skills by semantic relatedness, another task absent from training.
Moreover, the SkillMatch-1k evaluation data~\cite{decorte2024skillmatch} consists of skill mentions mined from job advertisements via lexical patterns, whose surface forms differ from the clean ESCO skill labels used as training anchors.

\niboldhead{Unseen ontology.}
We evaluate on two ontologies unseen during training: O*NET (v30.1)~\cite{tsacoumis2010onet} and the Singapore SkillsFuture (SSF) Skills Framework~\cite{skillsfuture}, with both unseen query and target distributions.
Table~\ref{tab_zero_shot_crossontology} reports results against the strongest baselines.
UWE outperforms the strongest baseline on O*NET and SSF by $+8.9$ and $+2.1$ MAP respectively, confirming that the learned representations generalize beyond the ESCO training distribution. %

\begin{table}[!ht]
\centering
\resizebox{\linewidth}{!}{%
\begin{tabular}{@{}l cc cc cc@{}}
\toprule
& \multicolumn{2}{c}{\textbf{Job2Skill}} & \multicolumn{2}{c}{\textbf{Skill2Job}} & \multicolumn{2}{c}{\textbf{Task Avg}} \\
\cmidrule(lr){2-3} \cmidrule(lr){4-5} \cmidrule(lr){6-7}
\textbf{Model} & MAP & \RPmini & MAP & \RPmini & MAP & \RPmini \\
\midrule
\multicolumn{7}{@{}l}{\textit{O*NET}} \\
\addlinespace[2pt]
UWE (Ours) & \textbf{32.9} & \textbf{39.3} & \textbf{37.2} & \textbf{57.0} & \textbf{35.1} & \textbf{48.1} \\
ContextMatch & 20.3 & 21.9 & 32.0 & 51.1 & 26.2 & 36.5 \\
Qwen3-8B & 18.0 & 20.7 & 33.3 & 53.4 & 25.7 & 37.0 \\
\addlinespace[4pt]
\multicolumn{7}{@{}l}{\textit{SSF}} \\
\addlinespace[2pt]
UWE (Ours) & \textbf{13.0} & \textbf{20.2} & \textbf{26.2} & \textbf{32.3} & \textbf{19.6} & \textbf{26.3} \\
ContextMatch & 11.1 & 18.4 & 22.2 & 28.4 & 16.7 & 23.4 \\
Qwen3-8B & 11.4 & 19.3 & 23.7 & 29.5 & 17.5 & 24.4 \\
\bottomrule
\end{tabular}%
}
\caption{Cross-ontology zero-shot results on O*NET and SkillsFuture (SSF). Neither ontology is seen during training. UWE outperforms the best specialist (ContextMatch) and generalist (Qwen3-8B) baselines.
\label{tab_zero_shot_crossontology}
}
\end{table}

\subsection{Gender-Stereotype Sensitivity}
\label{sec:bias_main}

As an initial diagnostic of bias in work-domain embeddings, we evaluate gender-stereotype associations using the Sentence Encoder Association Test (SEAT)~\cite{may2019seat}, comparing the pretrained MPNet backbone as initial model, to UWE and its late-interaction variants.
We assess two benchmarks: SEAT-C6b (male/female vs.\ career/family)~\cite{caliskan2017semantics} and a domain-specific test using ESCO occupation titles from engineering and creative arts professions.

Both MPNet and UWE exhibit statistically significant gender-occupation associations ($p < 0.01$) on both benchmarks, with UWE retaining similar effect sizes to its base model on SEAT-C6b and showing increased effect sizes on the ESCO benchmark.
This indicates that work-domain fine-tuning does not mitigate inherited stereotypes.
Late-interaction variants show comparable patterns.
Full methodology and detailed results are provided in Appendix~\ref{apdx:fairness}; limitations and recommendations are discussed in Section~\ref{sec:limitations}.

\section{Conclusion}
\label{sec_conclusion}
This work introduces a first unified framework for multi-task embeddings in labor market intelligence.
We contribute WorkBench, a unified evaluation suite spanning six work-domain tasks, and Unified Work Embeddings (UWE), a bi-encoder tackling all tasks with a single model.
Our analysis reveals significant cross-task transfer, with UWE outperforming the strongest specialist by $+2.6$ MAP and the best generalist (Qwen3-8B) by $+4.4$ MAP, while using two orders of magnitude fewer parameters and enabling zero-shot generalization.
\looseness=-1

\section{Limitations}
\label{sec:limitations}
\paragraph{Language scope.}
WorkBench and UWE are developed and evaluated exclusively on English data. While the underlying architecture and training methodology are language-agnostic, we intentionally focus on English to establish a rigorous multi-task foundation. This scope enables controlled experimentation and principled ablations without confounding effects from cross-lingual transfer. Nevertheless, real-world labor market applications are inherently multilingual. Hence, we encourage future work to extend WorkBench along the multilingual axis, leveraging multilingual encoders~\cite{jobbert_v3} and ontologies such as ESCO's multilingual structure~\cite{le2014esco}.

\paragraph{Model usage and bias mitigation.}
Given the breadth of evaluations and model contributions in this work, our gender-association analysis (Section~\ref{sec:bias_main} and Appendix~\ref{apdx:fairness}) is intentionally preliminary, serving as an initial diagnostic rather than a comprehensive audit. The SEAT-based evaluation measures intrinsic associative structure in the embedding space, but does not directly quantify downstream impact on hiring or matching outcomes. Therefore, we recommend task-specific audits before deployment in sensitive settings. 
We encourage future work to focus on extensive bias analysis across protected attributes in our proposed multi-task and zero-shot setup, and to investigate debiasing techniques tailored to multi-task work embeddings.

\clearpage
\appendix

\section{Data and Code Availability}
Upon acceptance, we will release the model, WorkBench evaluation suite, and code to facilitate reproducibility and future research.

\subsection{Data license overview}
In accordance with the ACL Responsible NLP Research Checklist, 
we summarize the licenses for third-party data sources in this work: 
\begin{itemize}
  \item \textbf{ESCO ontology}~\cite{le2014esco} is released under the Creative Commons Attribution 4.0 license (CC BY 4.0).
  \item \textbf{SkillExtract benchmarks} HOUSE
  and TECH
  subsets~\cite{decorte2022design} are derived from SkillSpan~\cite{zhang-etal-2022-skillspan} with ESCO skill annotations and are released under CC BY 4.0. 
  \item \textbf{SkillMatch-1K}~\cite{decorte2024skillmatch} used for SkillSim is released under the MIT license.
  \item \textbf{JobBERT evaluation dataset}~\cite{jobbert} used for JobNorm is released under the MIT license.
\end{itemize}
The released derivatives upon acceptance of this paper follow the applicable upstream terms and the licenses listed above.

\subsection{AI Writing and Coding Assistance}
During the preparation of this manuscript, the authors used generative AI tools, including ChatGPT, Gemini, and Claude, for language polishing and proofreading of author-written content. Additionally, Cursor (an AI-assisted coding environment) along with the aforementioned language models were used for coding assistance, specifically for debugging, code review, and implementation of routine components. In all cases, these tools served in an assistive capacity for short-term, well-defined tasks under direct human supervision. The authors retain full responsibility for the correctness and originality of the research ideas, methodology, experimental results, and all content presented in this work. No AI-generated text was used to produce novel research ideas or substantive written content.

\section{Preliminary Analysis of Gender-Occupation Associations}
\label{apdx:fairness}
\paragraph{Motivation and scope.}
Language models trained on large-scale corpora risk encoding societal biases that propagate to downstream applications~\cite{bolukbasi2016man}.
For work-domain models, such biases warrant particular scrutiny as systematic associations between demographic attributes and occupations or skills could adversely affect hiring recommendations and talent matching systems.

Given the breadth of evaluations and models studied in this work, a comprehensive bias analysis for multi-task work embeddings is beyond our scope.
Nevertheless, given the real-world implications of work-domain NLP, we report a preliminary sensitivity probe to characterize baseline bias properties.
We adopt the Sentence Encoder Association Test (SEAT)~\cite{may2019seat}, a standard intrinsic evaluation protocol for probing stereotypical associations in embedding spaces.
Since UWE inherits its initialization from MPNet and is adapted to synthetic data that may encode implicit biases, we expect the model to exhibit detectable stereotypical associations.
Based on this preliminary investigation, we encourage future work to focus on comprehensive bias auditing and mitigation for multi-task work embeddings.

\paragraph{Method.}
Following \citet{may2019seat}, SEAT defines two \emph{target} embedding sets $X$ and $Y$ of equal size (e.g., male vs.\ female) and two \emph{attribute} sets $A$ and $B$ (e.g., career vs.\ family). The test statistic is the difference of sums of the target concepts
\[
\begin{aligned}
s(X,Y,A,B)
&= \Bigl[ \displaystyle
   \textstyle\sum_{x \in X} s(x,A,B) \\
&\quad - \textstyle\sum_{y \in Y} s(y,A,B) \Bigr]
\end{aligned}
\]
where for the embeddings of an input sentence $q$, we compute an association score with the attributes
\[
s(q, A, B) = \mathbb{E}_{a \sim A}\big[\mathrm{sim}(q,a)\big] - \mathbb{E}_{b \sim B}\big[\mathrm{sim}(q,b)\big],
\]
with $\mathrm{sim}(\cdot,\cdot)$ the model's similarity function.
UWE computes similarity via its task-agnostic soft late interaction over token-level embeddings, while MPNet uses cosine similarity on mean-pooled sentence embeddings.
We perform the SEAT permutation test, reporting effect size (Cohen's $d$) and significance at $p < 0.01$.
Significant results indicate detectable bias; non-significant results should be interpreted as failure to detect bias rather than evidence of its absence.

\paragraph{Association tests.}
We evaluate two gender-association benchmarks.
The first, \textbf{SEAT-C6b (Male/Female vs.\ Career/Family)}, is a standard benchmark from \citet{caliskan2017semantics} adapted to sentence encoders by \citet{may2019seat}, measuring associations between gendered target sentences and career/family attributes.

To approximate downstream task conditions, we construct a second benchmark, \textbf{ESCO Engineering/Arts (M/F)}, using actual job titles from the work domain.
Resembling implicit human association in Caliskan Test 7 and 8~\cite{caliskan2017semantics}, we measure the association between male and female targets, and attributes engineering and arts.
Male and female target sentences are obtained from the SEAT-C6b test. Additionally, we collect job titles from ESCO~\cite{le2014esco}, selecting ISCO prefix 214 (Engineering professionals, excluding electrotechnology; 124 titles) and prefix 265 (Creative and performing artists; 56 titles), rebalancing to equal size via random subsampling.
Table~\ref{tab:targets_attributes_examples} shows representative examples.

\begin{table}[!t]
\centering
\resizebox{\linewidth}{!}{%
\begin{tabular}{@{}p{0.60\linewidth}p{0.62\linewidth}@{}}
\toprule
\textbf{Target concepts} & \textbf{Attribute concepts} \\ \midrule
\multicolumn{1}{@{}l}{\textit{Male}} & \multicolumn{1}{l}{\textit{Engineering}} \\
`The sons are there.', `A male is a person.', `Boys are people.',...
&
`engine designer', `food production engineer', `environmental expert',...
\\ \midrule
\multicolumn{1}{@{}l}{\textit{Female}} & \multicolumn{1}{l}{\textit{Arts}} \\
`The girls are there.', `Here she is.', `These are sisters.',...
&
`music director', `sculptor', `video artist',...%
\\
\bottomrule
\end{tabular}%
}
\caption{Example target and attribute concepts for ESCO Eng/Arts (M/F), showing three instances per category.}
\label{tab:targets_attributes_examples}
\end{table}

\paragraph{Results.}
Table~\ref{tab:fairness_uwe_mpnet} reports effect sizes for MPNet (UWE's base encoder) and UWE.
Both models exhibit statistically significant associations on both benchmarks, confirming gender-occupation stereotypes in the embedding space.
UWE retains similar effect size to MPNet on SEAT-C6b and increased effect size on ESCO Eng/Arts.

\begin{table}[!t]
\centering
\resizebox{\linewidth}{!}{%
\begin{tabular}{@{}lcc@{}}
\toprule
 & \textbf{ESCO Eng/Arts (M/F)} & \textbf{SEAT-C6b} \\
\midrule
MPNet (base model) & 1.160* & 0.388* \\
UWE & 1.570* & 0.392* \\
\bottomrule
\end{tabular}%
}
\caption{SEAT effect sizes for UWE and its base model.
$^{*}$indicates $p < 0.01$.}
\label{tab:fairness_uwe_mpnet}
\end{table}

As a secondary analysis, we investigate whether UWE's architectural design affects measured associations, specifically the introduced token-level late interaction.
Table~\ref{tab:fairness_lateinteraction} evaluates the late-interaction variants from the main paper's ablation study  in Section~\ref{sec:ablation_study}.
All variants show significant bias on ESCO Eng/Arts, while results on SEAT-C6b are more variable with sentence embeddings showing no significant effect.
These preliminary findings suggest that interaction mechanism influences measured associations, though the relationship warrants further investigation.

\begin{table}[!t]
\centering
\resizebox{\linewidth}{!}{%
\begin{tabular}{@{}lcc@{}}
\toprule
\textbf{Late Interaction} & \textbf{ESCO Eng/Arts (M/F)} & \textbf{SEAT-C6b} \\
\midrule
Sentence embeddings & 1.508* & 0.342 \\
$\operatorname{MaxSim}$ & 1.556* & 0.449* \\
$\operatorname{SoftMax}$-$y$-mean & 1.503* & 0.389* \\
$\operatorname{SoftMax}$ & 1.479* & 0.503* \\
\bottomrule
\end{tabular}%
}
\caption{SEAT effect sizes across late-interaction variants.
$^{*}$indicates $p < 0.01$.}
\label{tab:fairness_lateinteraction}
\end{table}

\paragraph{Limitations and future work.}
Given the breadth of evaluations and model contributions in this work, our bias analysis is necessarily preliminary, hoping to inspire future work to focus on extensive bias analysis in our proposed multi-task and zero-shot setup. This preliminary analysis is limited to gender; future work should examine additional sensitivity to protected attributes and social biases. Beyond analysis, we encourage future work to investigate whether established debiasing techniques transfer effectively to multi-task work embeddings.

\section{Synthetic Data Ablation: Full Results}
\label{apdx:synth_ablation_full}

Table~\ref{tab_synth_ablation_full} reports the per-task breakdown of the synthetic data ablation summarized in Section~\ref{sec:ablation_study}. Removing all synthetic enrichment (w/o all synth.), includes removal of synthetic aliases, synthetic job titles from inversed minority enrichment, and synthetic vacancy training data.
Alias removal alone (w/o alias synth.) has a mixed effect across tasks, with an average decrease of $-0.3$ MAP and $-0.6$ \RPmini.
Removing all synthetic data, results show this impacts performance of the related tasks directly for SkillNorm, JobNorm, and the skill extraction tasks. Additionally, the synthetic data helps to improve on other tasks such as Job2Skill, SkillSim, and with the largest improvement on Skill2Job.

\begin{table*}[!ht]
\centering
\resizebox{\textwidth}{!}{%
\setlength{\tabcolsep}{5pt}%
\begin{tabular}{@{}lcccccccccccccccc@{}}
\toprule
& \multicolumn{2}{c}{\textbf{Job2Skill}} & \multicolumn{2}{c}{\textbf{Skill2Job}} & \multicolumn{2}{c}{\textbf{JobNorm}} & \multicolumn{2}{c}{\textbf{SkillNorm}} & \multicolumn{2}{c}{\textbf{SkillSim}} & \multicolumn{2}{c}{\textbf{SkillExtr-T}} & \multicolumn{2}{c}{\textbf{SkillExtr-H}} & \multicolumn{2}{c}{\textbf{Task Avg}} \\
& MAP & \RPmini & MAP & \RPmini & MAP & \RPmini & MAP & \RPmini & MAP & \RPmini & MAP & \RPmini & MAP & \RPmini & MAP & \RPmini \\
\midrule
\textbf{UWE} & \textbf{17.9} & \textbf{37.8} & \textbf{37.3} & \textbf{49.4} & \text{39.3} & \textbf{60.8} & \textbf{88.9} & \textbf{95.9} & \textbf{10.4} & \textbf{33.8} & \textbf{52.3} & \textbf{73.5} & \text{42.3} & \textbf{64.2} & \textbf{40.2} & \textbf{57.7} \\
w/o alias synth. & 17.7 & 37.5 & 37.0 & 49.2 & \textbf{39.6} & 60.7 & 87.9 & 95.3 & 9.9 & 32.3 & 51.8 & 73.3 & \textbf{42.6} & 62.4 & 39.9 & 57.1 \\
w/o all synth. & 17.7 & 37.6 & 35.9 & 47.7 & 37.7 & 57.4 & 87.7 & 94.9 & 10.3 & 32.4 & 46.0 & 69.6 & 37.5 & 61.2 & 38.5 & 55.9 \\
\bottomrule
\end{tabular}%
}
\caption{Full per-task results of the synthetic data ablation on WorkBench, reporting MAP and \RPmini{}. ``w/o alias synth.'' removes synthetic skill aliases; ``w/o all synth.'' removes all synthetic enrichment in the training data.
\label{tab_synth_ablation_full}}
\end{table*}

\section{Experiment Setup Details}  %
\label{apdx:exp_details}

\subsection{Hyper-parameter Search}\label{apdx:gridsearch}
To obtain an experimental configuration, we first perform a gridsearch of the hyper-parameters, evaluated on task-average MAP performance after 600 update steps. We denote the best setting in bold in the following.
\begin{itemize}
    \item \textbf{Soft Late Interaction temperature} $\tau_a$ is the temperature used for scaling of the $\operatorname{SoftMax}$ to obtain the interaction matrix $A$. We perform a gridsearch over $\{ 2.0, 1.0, \textbf{0.5}, 0.1\}$.
    \item \textbf{Per-task loss components} are $\alpha_J, \alpha_V$, and $\alpha_A$. For a gridsearch over values $\{ 0, 0.5, 1.0\}$, we find configuration $\alpha_J=1.0$, $\alpha_V=0.5$, $\alpha_A=0.5$.
    \item \textbf{Learning rate} for the AdamW optimizer is $\{1 \times 10^{-4}, \mathbf{8\times 10^{-4}}, 5\times 10^{-5}, 1\times 10^{-5}\}$.
    \item\textbf{MTM loss temperature $\tau$} in Eq.~\ref{eq_asym_loss} has a gridsearch over $\{0.01, 0.02, 0.05, 0.1\}$ for the 3 loss components, resulting in $0.05$ for $\mathcal{L}_{S,J}$, and $0.02$ for $\mathcal{L}_{S,V}$, $\mathcal{L}_{S,A}$.
\end{itemize}

\noindent The following configurations are kept constant:
\begin{itemize}
  \item \textbf{Base encoder:} \texttt{all-mpnet-base-v2} (109M parameters)\footnote{\scriptsize
    \url{https://huggingface.co/sentence-transformers/all-mpnet-base-v2}
  }%
    , a state-of-the-art MPNet-based model~\cite{song2020mpnet}.
  \item \textbf{Batch composition:} \mbox{$512$} skills, $512$ job titles,
        $512$ vacancy sentences, $512$ skill alternatives
        ($2{,}048$ texts per update step).
  \item \textbf{Optimiser and learning rate schedule:} We use the AdamW optimizer~\cite{adamw} with a learn learning rate schedule where warm-up uses the first $10\%$ of training length, followed by linear decay. Peak learning rate is defined by the gridsearch above.
  \item \textbf{Token budget:} To constrain memory size during training, we set a max token capacity of $64$ for all inputs, with no constraints during inference. We find this budget sufficient to support the sentence length of all WorkBench tasks
  \item \textbf{Hardware:} All experiments are conducted on a NVIDIA A100‑SXM4‑40GB GPU.
\end{itemize}

\subsection{Experiment Details}
\subsubsection{Loss ablation.} Results in Table~\ref{tab_loss_ablation} of the main paper use the best setup defined in the hyper-parameter search. However, to disentangle the effects of using our soft late interaction, we apply the standard mean pooling of the embeddings instead. The final results are averaged over 5 runs over different seeds of $600$ update steps each, over which we report the $95\%$ confidence interval.

\subsubsection{Late interaction ablation.} 
\begin{figure}[!b]
  \centering
    \includegraphics[width=1.0\columnwidth,trim=0pt 3pt 5pt 5pt,clip]{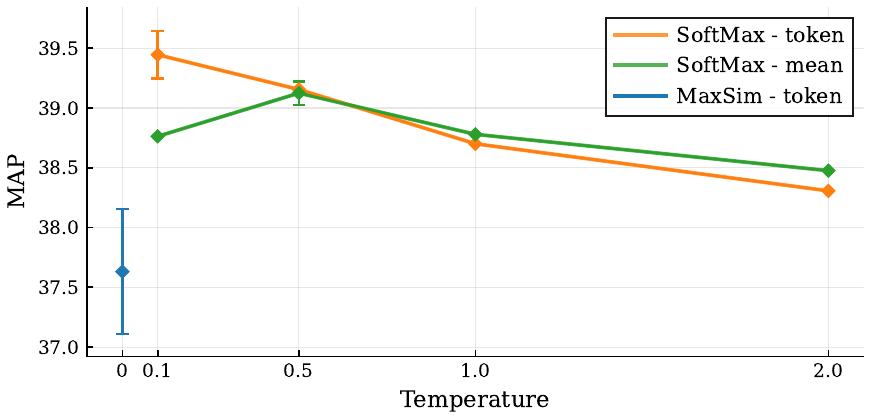}
  \caption{Overview of temperature $\tau_a$ influence for late interaction, including SoftMax with token-level embeddings (token), and target-averaged embeddings (mean). MaxSim is depicted as having temperature zero. The final results after the gridsearch are depicted with $95\%$ CI.
  }
  \label{fig_apdx_late_interaction_ablate}
\end{figure}

Results in Table~\ref{tab_LI_numbers} of the main paper also use the best setup defined in the hyper-parameter search. Similar to the previous setup, the final results are averaged over 5 runs over different seeds of $600$ update steps each, over which we report the $95\%$ confidence interval.
Figure~\ref{fig_apdx_late_interaction_ablate} shows an overview of influence of the temperature $\tau_a$ on the {SoftMax} objectives, where {MaxSim} is depicted as having $\tau_a=0$.
The results show that the token-level Softmax outperforms the target-averaged version, but only for a lower temperature. After the gridsearch, the best setting is run for the 5 seeds, resulting in $\tau_a=0.1$ for token-level SoftMax (\emph{Softmax - token}), and $\tau_a=0.5$ for target-averaged embeddings (\emph{Softmax - mean}).

\subsubsection{Cross-task transfer.} To obtain results in Figure~\ref{fig_heatmap_task_transfer} in the main paper, we make a reduced $300$ optimisation steps per objective due to faster convergence of the single-component losses. The experiment uses the best setup defined in the hyper-parameter search. For each of  $\mathcal{L}_{S,J}$, $\mathcal{L}_{S,V}$, $\mathcal{L}_{S,A}$, we make 5 runs over different seeds and report standard error (SE). We calculate the Knowledge Gain based on the average per-task performance of the final model and the base MPNet model from which training was initiated.

\subsubsection{Timing evaluation.} To obtain the latency and FLOPS per query in Table~\ref{tab_latencies} in the main paper, we run each baseline in isolation. To measure the metrics at inference, we cache the model target space embeddings, and make a measurement for each query in a subset per task. 
We enable a warmup of 30 queries to avoid additional latencies on initial memory allocations, then we measure 100 queries per task to measure latencies, and 10 separate queries per task to measure the FLOPS. We allocate more budget to latencies due to its high variability. The final reported results first average per task over all queries, and then calculate the macro-average on WorkBench with standard error over tasks.

\subsubsection{Final WorkBench evaluation.} The results in Table~\ref{tab_main_workbench_results} in the main paper rely on the publicly available models for each of the baselines. The task-specific models JobBERT-v2 and ContextMatch are additionally verified to have matching results from their original JobNorm, and SkillExtr-H and SkillExtr-T
 benchmarks~\cite{decorte2025contextmatch_jobbertv2}.
 For our final model, we rely on the hyperparameters as described in the gridsearch above. Furthermore, following \citet{decorte2025contextmatch_jobbertv2}, we randomly apply a prefix or suffix vacancy sentence to each of the vacancy sentences. We ensure it is a non-matching sentence with any vacancy sentence in the mini-batch, by sampling a vacancy sentence for the out-of-batch skills. We ablate the sampling probability and apply a gridsearch over values of $\{0.0, \textbf{0.8}\}$.
 In contrast to the ablation experiments, we evaluate the model every 50 update steps and retain the final model with the best task-average evaluation performance.

\section{Data pipeline details}
\label{apdx:data_pipeline_details}

\subsection{Inversed synthetic minority enrichment.} 

\subsubsection{Synthetic job title generation}
To obtain the generated job titles for minority skills, we generate job titles with a generative language model. Using GPT4o-mini~\cite{achiam2023gpt}, we set a minimum number of $N=20$ generations for each ESCO skill. This ensures all minority skills are represented by a minimal subset of job titles. Additionally, this ensures majority skills are enriched from a similar data-generating distribution as the minority skills. Using structured output, we enforce a list-like return type of the generative model. However, we found this resulted in a  wide range in the number of generations per skill. Especially as we parse the results to be valid only if the generated title contains at least two words.
Therefore, we enabled a retry mechanism that would retry the prompt asking for the remaining number of required job titles, and providing the already generated job titles in context.
Instead of generating a single job title $N$ times independently, this procedure ensures we have a minimal of duplicate job titles per skill by providing context to the generative model.

The prompt is formulated as follows  for an initial $N=20$ job titles to be generated, per skill $s_i$, with ESCO description $d_i$:
\texttt{
    Return a list of $N$ unique, real-world job titles for which the given skill is required. 'Real-world' means titles actually used in industry or job ads. Make heavy use of alternative spellings, paraphrases, synonyms. Add heavy noise such as alternative upper casing, spellings, paraphrases, synonyms. Add the noisy information using comma-separation, brackets, \&, @, -, or other separators.  Remove exact duplicates so every title is unique. The skill can only occur explicitly in the first job title. You are given: <skill\_name>\textbf{$s_i$}</skill\_name> <skill\_description>\textbf{$d_i$}</skill\_description>
}

\noindent When a retry is applied since there are insufficiently generated job titles for the skill, $N$ is set to the remaining number of required job titles, and the prompt is appended with the already generated job titles $J$ (variables indicated in bold):
\texttt{
You already provided these job titles, make sure there are no duplicates: \textbf{$J$}
}

\subsubsection{Merging synthetic job titles into real-world data.} After generating job titles, Section~\ref{sec_method} describes the merging procedure by matching synthetic job titles with existing real-world job titles in the dataset. In this way, the real-world job title skill profiles get enriched with minority skills. We describe the job merging in more detail in the following. 

\begin{figure}[t]
  \centering
    \includegraphics[width=1.0\columnwidth,trim=0pt 3pt 5pt 30pt,clip]{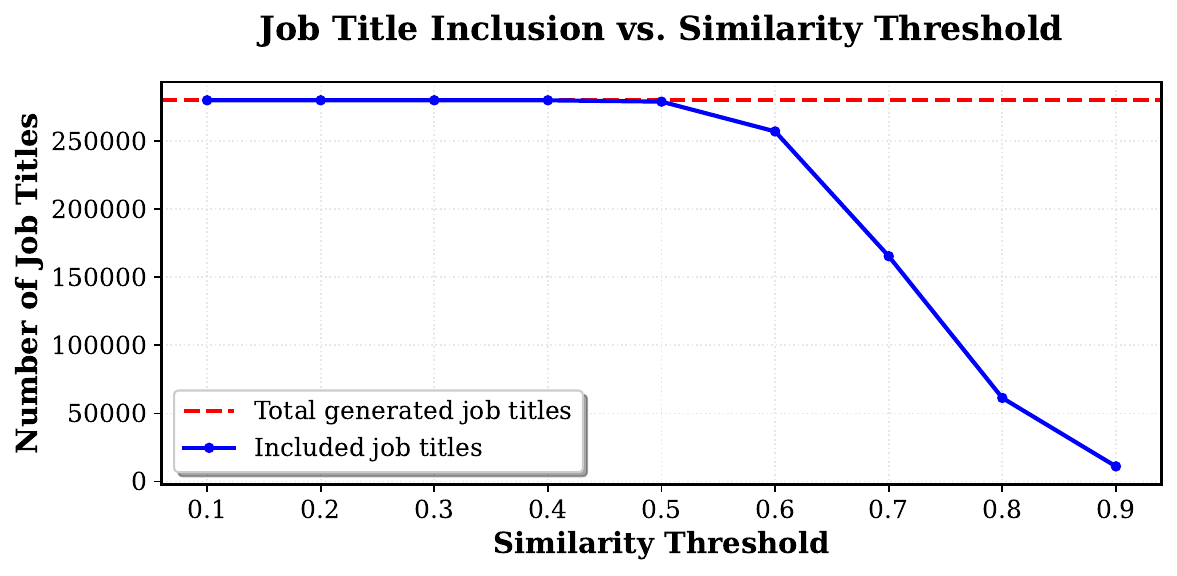}
  \caption{Job title similarity threshold over the number of included jobs. Synthetic job title skills are merged with real-world job titles when above the similarity threshold.  
  \label{fig_apdx_jobnorm_threshold}
  }
  
\end{figure}

Using state-of-the-art model JobBERT-v2~\cite{decorte2025contextmatch_jobbertv2} we find the job similarity between the synthetic and real-world job titles. A match is determined based on a similarity threshold of the cosine similarity between the embeddings of both job titles. 
Figure~\ref{fig_apdx_jobnorm_threshold} depicts the total number of generated job titles, and illustrates the included job titles for a range of thresholds. After qualitative evaluation on a subset, we pick the threshold $0.7$ that results in $165$k out of the $280$k generated titles having a match in the real-world job titles. A synthetic job title can have multiple matches with real-world job titles, as long as the similarity is above the threshold. While this work focuses on English, the procedure is extensible beyond an English setting using a multilingual encoder~\cite{jobbert_v3}.

\subsection{Fully synthetic data for SkillNorm.}
For skill normalization, WorkBench employs the full ESCO dataset for validation and test sets.
For training data, however, this task is specifically suited for synthetic data generation.
To this end, we employ GPT4o-mini~\cite{achiam2023gpt} to generate a set of alternative formulations $A$ for each of the ESCO skills. 
The prompt for skill $s_i$ is formulated as follows, including a fixed set of 3 representative examples from the validation set:
\texttt{Return a LIST OF 20 UNIQUE PARAPHRASES of the given skill. Each paraphrase should express the same core meaning but use different wording. Use synonyms, change word order, add or modify verbs, make it more specific. Examples: supervise correctional procedures -> oversee prison procedures,  show initiative -> initiate action lead police investigations -> manage police enquiries. Remove exact duplicates so every paraphrase is unique. Maintain the same skill meaning while varying the expression. You are given: <skill\_name>\textbf{$s_i$}</skill\_name>}
\looseness= -1

\end{document}